\documentclass{article}

\usepackage[final]{corl_2020} 
\usepackage{graphicx,amsmath,amssymb,array,fancyhdr,color,sidecap,paralist,eurosym,nicefrac,lastpage, textcomp}
\usepackage{tikz}
\usepackage{bm}
\usepackage{booktabs}
\usepackage[percent]{overpic}
\usepackage{hyphenat}
\usepackage{color}
\usepackage{amsmath}
\usepackage[linesnumbered,ruled]{algorithm2e}
\usepackage{subcaption}
\usepackage{bm}
\newcommand{\ppi}{\bm{\pi}_{\bm{\theta}}}
\newcommand{\pf}{p}
\newcommand{\pobs}{p}
\newcommand{\psamples}{{q_{\mathrm{samples}}}}
\newcommand{\pdata}{{q_{\mathrm{data}}}}
\newcommand{\pdatapi}{{q_{\mathrm{\bm{\theta}, data}}}}
\newcommand{\pds}{{q}}
\newcommand{\st}{{\bm{\xi}}}
\newcommand{\obs}{{\bm{y}}}
\newcommand{\q}{{\bm{q}}}
\newcommand{\nbdata}{N}
\newcommand{\dq}{{\dot{\bm{q}}}}
\newcommand{\ddq}{{\ddot{\bm{q}}}}
\newcommand{\x}{{\bm{x}}}

\newcommand{\tar}{\bm{d}}
\newcommand{\dx}{\dot{\bm{x}}}
\newcommand{\ddx}{\ddot{\bm{x}}}
\newcommand{\ic}{{\bm{u}}}
\newcommand{\traj}[1]{\bm{\tau}^{#1}}
\newcommand{\nn}{\mathrm{F}}
\newcommand{\trq}{\bm{\tau}}
\newcommand{\force}{\text{\textbf{f}}}
\newcommand{\ppm}{{\bm{\theta}}}
\newcommand{\dpm}{{\bm{\theta}_f}}
\newcommand{\mup}{{\bm{\mu}}}

\newcommand{\J}{{\bm{J}}}
\newcommand{\Sigmap}{{\bm{\Sigma}}}

\newcommand{\pmvn}{\mathcal{N}}

\newcommand{\unit}[1]{~\mathrm{#1}}

\newcommand{\trans}[2]{\mathcal{T}_{#2}^{#1}}
\newcommand{\transs}[2]{\mathcal{T}_{\bm{\xi},#2}^{#1}}
\newcommand{\transu}[2]{\mathcal{T}_{\bm{u}, #2}^{#1}}

\newcommand{\trsp}{{\scriptscriptstyle\top}}

\title{Generative adversarial training of product of policies for robust and adaptive movement primitives}

\author{
 Emmanuel Pignat\\
  Idiap Research Institute and EPFL\\
  Switzerland\\
  \texttt{emmanuel.pignat@gmail.com} \\
  \And
  Hakan Girgin\\
  Idiap Research Institute and EPFL\\
  Switzerland\\
  \texttt{hakan.girgin@idiap.ch} \\
  \And
  Sylvain Calinon\\
  Idiap Research Institute and EPFL\\
  Switzerland\\
  \texttt{sylvain.calinont@idiap.ch} \\
}

\begin{document}
\maketitle


\begin{abstract}
    In learning from demonstrations, many generative models of trajectories make simplifying assumptions of independence. Correctness is sacrificed in the name of tractability and speed of the learning phase. 
    The ignored dependencies, which often are the kinematic and dynamic constraints of the system, are then only restored when synthesizing the motion, which introduces possibly heavy distortions. 
    In this work, we propose to use those approximate trajectory distributions as close-to-optimal discriminators in the popular generative adversarial framework to stabilize and accelerate the learning procedure. 
    The two problems of adaptability and robustness are addressed with our method. 
    In order to adapt the motions to varying contexts, we propose to use a product of Gaussian policies defined in several parametrized task spaces. Robustness to perturbations and varying dynamics is ensured with the use of stochastic gradient descent and ensemble methods to learn the stochastic dynamics. Two experiments are performed on a 7-DoF manipulator to validate the approach.
\end{abstract}


\section{Introduction}

Adaptability and ease of programming are key features necessary for a wider spread of robotics in factories and everyday assistance.
Learning from demonstrations (LfD) is an approach to address this problem. 
It aims to develop algorithms and interfaces such that a non-expert user can teach the robot new tasks by showing examples.  
In LfD, movements are commonly represented using movement primitives (MPs). They are used as building blocks of more complete skills in which they can be combined sequentially or simultaneously.

In LfD, the parameters of MPs are learned from a set of demonstrations. Ideally, motions synthesized from the MPs should match the distribution of demonstrations with the same variability \cite{paraschos2013probabilistic, englert2013probabilistic}. It can be later exploited for multiple usages, such as including additional constraints or objectives. Furthermore, keeping the variability is primordial when the demonstrations are used to initialize policy search \cite{levine2013guided}. 
Another desired feature of MPs is their adaptation to new situations or targets, such as moving objects. To that end, a common approach is to learn MPs in multiple parametric task spaces\footnote{In this work, task spaces are not limited to the position and orientation of the end-effector but are general transformations of the configuration space. The configuration space itself will be considered as a task space with an identity transformation.} \cite{calinon2016tutorial, calinon2009statistical, paraschos2017probabilistic}. For example, the task spaces can be attached to objects of interest \cite{niekum2015learning, muhlig2009automatic, calinon2016tutorial} such that the movements are analyzed under several coordinate systems. However, for computational reasons, many of these approaches make an assumption of independence between the MPs in the different spaces; those are learned independently and only combined at the controller level, which results in distortions in the synthesis. 

A consistent framework for learning multiple models jointly is the product of experts (PoE) \cite{hinton2002training, zen2012product}. 
Models with unnormalized likelihood like PoEs are used in robotics for inverse optimal control (IOC) \cite{ziebart2008maximum, finn2016guided, kalakrishnan2013learning}. However, they either rely on expensive approximations of the normalizing constant \cite{finn2016guided} or learn only weights of predefined features \cite{kalakrishnan2013learning}. 
Generative adversarial modelling \cite{goodfellow2014generative} has been proposed as a more efficient approach with improved training stability \cite{finn2016connection, hausman2017multi, ho2016generative}.

In this work, we propose to train MPs within the generative adversarial framework which we call generative adversarial movement primitives (GAMP). We propose several adaptations of the discriminator and a particular parametrization of the policy to meet the requirements of LfD; as performing demonstrations on the physical systems is costly, we typically only have a few trajectories (from 5 to 20 depending on the complexity). Also, the training process should be interactive and thus relatively fast (from a few seconds to a few minutes). 
The proposed approach can be classified as model-based imitation learning \cite{englert2013probabilistic}. Given a prior knowledge about the dynamics of the system, it uses model-based policy search to minimize an imitation cost. As we will show through real robot experiments, rough dynamic models make the process sample-efficient. We also propose a variant of the method to refine these models through executions on the real system. Our approach treats both epistemic uncertainty (coming from partial knowledge of the system) and aleatoric uncertainty (coming from stochasticity of the system), resulting in robust controllers.
Finally, our framework aims to remain general and be compatible with multiple control strategies such as velocity, acceleration or torque control.
It can also be used to train both time-dependent \cite{paraschos2013probabilistic, calinon2009statistical} or time-independent policies \cite{khansari2011learning}.

Python/TensorFlow codes related to this paper can be cloned from the following repository:  \href{https://github.com/emmanuelpignat/tf_robot_learning}{https://github.com/emmanuelpignat/tf\_robot\_learning}.


\section{Generative adversarial training for product of policies}
\label{sec:gamp}
In the generative adversarial framework \cite{goodfellow2014generative}, a generator $G(\bm{z}; \bm{\theta})$ is trained to transform input noise $p_z(\bm{z})$ into samples that look like the data distribution. To do this, a discriminator is trained in parallel to output the probability that a sample comes from the data rather than from the generator. On its side, the generator has to maximize the probability to mislead the discriminator.
The generator and discriminator are typically neural networks trained with stochastic gradient descent (SGD). At each step of the training, the discriminator is optimized for a few steps of SGD and then one step is done for the generator. 

\subsection{State-space generator}
In order to generate trajectories, the considered generator is a state-space model defined by several components. We assume that we have access to a stochastic dynamic model of the robot
$\pf(\st_{t+1}\vert\st_t, \ic_t, \dpm)$
where $\st_t$ is the state at time t, $\ic_t$ the control command and $\dpm$ the parameters of the dynamics model. If the robot is controlled with inverse dynamics, this model can be a simple integrator, but more complex models such as neural networks can be considered.
If the dynamics model is not known or is uncertain, a distribution of parameters $p(\dpm)$ can be defined, under which the expected objective will be optimized, as we will see Sec.~\ref{sec:stochastic}. 
The state of the system being sometimes not directly observed, an additional component that needs to be defined is a stochastic observation model $\pobs(\obs_t | \st_t)$ where $\obs_t$ is an observation. Control commands are computed given this observation by a stochastic policy
$\ppi(\ic_t | \obs_t)$ 
where $\ppm$ are the parameters of the policy. 
A distribution of trajectories $\traj{} = \{ \obs_1, \ic_1, \dots, \obs_T, \ic_T\}$ is then defined as a state-space model with
\begin{align}
p(\traj{} | \dpm, \ppm) = 
p(\st_1) \prod_{t=1}^{T}p(\st_{t+1}|\st_t, \ic_t, \dpm)\, \pi_\ppm(\ic_t | \obs_t)\,
\pobs(\obs_t | \st_t).
\label{equ:ptraj}
\end{align}
An evident way to learn the MPs is to compute maximum likelihood estimation of $\ppm$ given a set of demonstrated trajectories.
However, computing or maximizing this likelihood requires approximations or restricting assumptions \cite{doerr2018probabilistic, deisenroth2011pilco, englert2013probabilistic}. 
In GANs, the likelihood is not modelled explicitly. It is just required to be able to draw samples from this density. Full sequences can be generated by forward sampling, by sampling each model after the other according to \eqref{equ:ptraj}. Thus, great flexibility is allowed for setting the dynamics, policy and observation models; 
they only have to be simple to sample from. Additionally, this process should be differentiable with respect to their respective parameters (e.g. $\ppm$ and $\obs$ for the policy model) using the reparametrization trick \cite{kingma2013auto}.
If the observation model is not bijective, it might be impossible to retrieve the distribution of initial states of the demonstrations $p(\st_1)$. In this case, this distribution can be parametrized and optimized as well. For simplicity of the notation, an observable system with $\st_t=\obs_t$ will be used for the derivations in the rest of the paper, without loss of generality. 

\subsubsection{Adaptation with products of Gaussian policies (PoGP)}
\label{subsec:pogp}
While the $d_\st$-dimensional state $\st$ of a robotic manipulator is defined by its joint angles (and possibly velocities), movements are often best explained under several task spaces. Each task space $P$ is associated with a task map, which is a non-linear function $\transs{}{p} : \mathbb{R}^{d_\st} \rightarrow  \mathbb{R}^{k_\st^p}$.  Accordingly, a set of linear functions maps control commands $\ic$ (joint velocities or acceleration) to their value in the different task spaces $\transu{}{p} : \mathbb{R}^{d_\ic} \rightarrow  \mathbb{R}^{k_{\ic}^p}$. 
These transformations can be parametrized by the poses of an external object (which we will drop in the notation for simplicity) or by time, which we will denote with the superscript $t$.
We propose to define a stochastic Gaussian policy in each of these task spaces as
\begin{align}
\transu{t}{p}(\ic_t) \sim \pmvn\Big(\mup_p^t\big(\transs{t}{p}(\st_t)\big), \Sigmap_p^t\big(\transs{t}{p}(\st_t)\big)\Big).
\end{align}
In the most general case, $\mup_p^t(\cdot)$ and $\Sigmap_p^t(\cdot)$ can be neural networks. In simpler cases, the policies can be proportional-derivative controllers with a constant covariance. In Appendix \ref{sec:appendix_policies}, different parameterizations of the policy are given.
The proposed overall policy is the fusion of the policies in the different task spaces, given as a product of linearly transformed Gaussians
\begin{align}
\pi_\ppm(\ic_t | \st_t) \propto \prod_{p=1}^{P} \pmvn\Big( \transu{t}{p}(\ic_t)\Big|\mup_p^t\big(\transs{t}{p}(\st_t)\big), \Sigmap_p^t\big(\transs{t}{p}(\st_t)\big)\Big).
\end{align}
This product has a closed form expression as a Gaussian and can be sampled directly.
Many works \cite{calinon2009statistical, paraschos2017probabilistic, pignat2019bayesian} have a final controller of this form, but it is computed by using expert policies that have been learned  independently.
\subsection{Including LfD generative models as close-to-optimal discriminators}
\label{subsec:discriminator}
Besides training a discriminator as a neural network $D(\traj{})$, we propose to include standard generative models used in LfD. They are usually trained in closed form or with very efficient procedures like EM. 
This addition is motivated by a dramatically increased stability and speed of the training procedure. 
We propose to include a second discriminator $D_\pds(\traj{})$ which is multiplied to the original one. This discriminator consists of two approximate distributions $\psamples$ and $\pdata$ learned with standard LfD generative models as \cite{paraschos2013probabilistic, calinon2009statistical, calinon2016tutorial}.
In this work, this discriminator is called a \textit{close-to-optimal discriminator} with
\begin{align}
D_\pds(\traj{}) = \frac{\pdata(\traj{})}{\pdata(\traj{}) + \psamples(\traj{})}.
\end{align}
In an optimal discriminator, $\pdata$ and $\psamples$ would be the exact distributions, not the approximate ones which are improperly normalized. But if we were able to model explicitly this likelihood, the generative adversarial approach would not be needed. The class of distributions $\pds$ we propose to use typically drops some dependencies or do not integrate to one on the space of trajectories. Another formulation of this problem is that the system is underactuated\footnote{All the trajectories of the distribution are not feasible.} \cite{osa2018algorithmic}. Directly used as generative models, where the dropped dependencies are only restored at the synthesis phase \cite{calinon2016tutorial, paraschos2017robot}, these models induce distortions, as discussed in \cite{zen2012product}. 
These distortions are extensively reduced if these models are used as classifiers in the context of generative adversarial learning; both the samples from the generator and the dataset are compared under the same approximations while the feasibility and dependencies are ensured by the generator \eqref{equ:ptraj}.

We propose to train the approximate distribution $\pdata$ once at the beginning of the learning process. 
The approximate distribution of samples $\psamples$ is updated with maximum likelihood before each step of gradient descent of the policy parameters, see Alg.~\ref{alg:adaptvitiy_gamp}. 
As it might be costly to generate many samples from the generator at each iteration, $\psamples$ can be learned incrementally with stochastic updates of maximum likelihood. Such updates can be derived for expectation maximization (EM), closed-form maximum likelihood (e.g.\ Gaussian distribution) or variational inference \cite{hoffman2013stochastic} (in the case where $\pdata$ and $\psamples$ are Bayesian models whose posterior distribution is estimated). 


Many possibilities are offered for choosing the family of approximate distributions $\pds$. A very simple choice, if the trajectories are all aligned in time, is to use a factorized Gaussian distribution as
\begin{align}
\small
\pds(\traj{}) = \prod_{t=1}^{T}\pmvn\Big(\begin{bmatrix}
\st_t\\\ic_t
\end{bmatrix} \Big|\, \mup_t, \Sigmap_t\Big).
\label{equ:q_gaussian}
\end{align}
Matching factorized Gaussian distributions is also done in \cite{englert2013probabilistic} in a similar context. However, $\pds$ is represented explicitly by using Gaussian process dynamics models and moment matching approximations \cite{deisenroth2011pilco}.
If the trajectories have particular correlations across time (that are not due to the dynamics), probabilistic movements primitives (ProMP)\cite{paraschos2013probabilistic} can be used instead.
In order to provide adaptation to parametrized task spaces, the discriminator can additionally compare the trajectories in these task spaces as 
\begin{align}
\small
\pds(\traj{}) = \prod_{t=0}^{T}\Bigg(\pmvn\Big(\begin{bmatrix}
\st_t\\\ic_t
\end{bmatrix}\Big|\, \mup_t, \Sigmap_t\Big)\prod_{p=1}^{P}\pmvn\Big(
\begin{bmatrix}
\transs{t}{p}(\st_t)\\
\transu{t}{p}(\ic_t)
\end{bmatrix}
\Big|\, \mup_t^p, \Sigmap_t^p\Big)\Bigg).
\label{equ:q_gaussian_task_space}
\end{align}
In this case, even if the approximate distributions $\pds$ are Gaussians, the discriminator would be able to distinguish between more complex distributions, as the comparison is done under non-linear transformations. 
In the case where the trajectories cannot be time-aligned or that the targeted distribution is multimodal, more complex models as hidden Markov models \cite{calinon2016tutorial} or Gaussian mixture models \cite{pignat2019bayesian} can be used, with the density
\begin{align}
\small
\pds(\traj{}) = \prod_{t=0}^{T}\Bigg(\sum_{k=0}^{K} \pi_k\ \pmvn\Big(\begin{bmatrix}
\st_t\\\ic_t
\end{bmatrix} \Big|\, \mup_k, \Sigmap_k\Big)\Bigg).
\label{equ:q_mixture}
\end{align}
They can be trained very efficiently with EM in a few milliseconds.

\begin{algorithm}[h]
	\small
	Compute maximum likelihood of $\pdata$ on the $\nbdata$ demonstrations $\{\hat{\traj{}}^{(i)}\}_{i=1}^\nbdata$ \\
	\For{number of training iterations}{
		\For{$L$ dynamic models $\{\dpm^{(j)}\}_{j=1}^{L}$}
		{
			Sample from \eqref{equ:ptraj} $M$ trajectories $\{\traj{(i, j)}\}_{i=1}^M$\\
			Apply (stochastic) maximum likelihood update on $\psamples^{(j)}$ given $\{\traj{(i, j)}\}_{i=1}^M$\\
		}
		Update \textbf{global} policy parameters $\ppm$ by descending the stochastic gradient:
		\tiny
		\begin{align}
		\nabla_\ppm \frac{1}{ML}\sum_{j=1}^{L}\sum_{i=1}^{M} \Big(-\log\big(\pdata(\traj{(i, j)})\big)+
		\log\big(\pdata(\traj{(i, j)}) + \psamples^{(j)}(\traj{(i, j)})\big) - \log\big(D(\traj{(i, j)})\big) \Big)
		\end{align}
	\vspace{-1em}	
}
	\caption{Robust generative adversarial training of movement primitives}
	\label{alg:adaptvitiy_gamp}
\end{algorithm}
By using Gaussian models, which have quadratic log-likelihood, the gradients are well-behaved, leading to fast convergence. At initialization, our generator samples trajectories $\traj{(i)}$ very far from $\pdata$ but close from $\psamples$. The term $\frac{1}{M}\sum_{i=1}^{M}-\log\big(\pdata(\traj{(i)})\big)$ in Alg.~\ref{alg:adaptvitiy_gamp} would dominate the gradient of the cost, which would be close to quadratic. For practical and stability reasons, we propose to train the policy first by considering only the additional classifier using approximate distributions. Then, the neural network classifier is only used for additional refinements and with smaller learning rates. In this case, the neural network is just helping the additional classifier to distinguish features that are not encoded in the approximate distributions $\pds$.
\section{Robustness and unknown dynamics}
\label{sec:stochastic}
In this section, we propose a strategy to learn robust policies in case of changing or unknown dynamics. So far, we have considered fixed parameters $\dpm$ of the stochastic dynamic system. It can result in a poor matching of the trajectory distribution in the case where the model of dynamics $\dpm$ does not match the real system. In the worst case, the distributions can completely diverge and executing the policy can be dangerous. In other cases, the problems can be less disastrous and more subtle. For example, if the model overestimates the stochasticity of the system, the rollouts on the real system would have lower variance than the demonstrations. In this case, the system will rely too much on the stochasticity of the environment to create variability.
A robust policy has to match the distribution of trajectories for a distribution of parameters $p(\dpm)$. This distribution can be either a hand-tuned prior distribution, a posterior distribution if the dynamics are learned with Bayesian methods or a set of parameters $\{\dpm^{(j)}\}_{j=1}^{L}$ if they are learned with ensemble methods.

We propose to condition the discriminator on $\dpm$, which means that this value should be fed to it together with the samples. As the true system is not known when executing the policy, this latter should not depend on the parameters of the dynamics. Giving access to the model parameters on which are generated the samples to the discriminator only forces the policy to match the distribution of data under a distribution of dynamic parameters, ensuring robustness.

In order to use the additional discriminators  $D_\pds(\traj{})$ proposed in Section \ref{subsec:discriminator}, two alternatives are possible. The choice mainly depends on the trade-off between robustness and computation time. In both cases, multiple approximate models $\{\psamples^{(j)}\}_{j=1}^{L}$ are learned on a batch of dynamic parameters $\{\dpm^{(j)}\}_{j=1}^{L}$. In the case of privileging robustness, the $L$ dynamic parameters are drawn from their distribution before each iteration of gradient descent. In this case, enough samples should be drawn from \eqref{equ:ptraj} in order to compute the maximum likelihood of the approximate distribution $\psamples^{(j)}$. The stochastic updates are not allowed as the $L$ model from the previous iterations do not correspond anymore. When it is too costly to sample enough trajectories to perform complete maximum likelihood of $\pds$, the $L$ models can be changed only after a given number of iterations or even kept fixed throughout the learning process. This solution is also natural if the parameters are learned by an ensemble method. The procedure is more formally presented in Alg.~\ref{alg:adaptvitiy_gamp}.

The parameters of the system $\dpm$ can be also learned or refined. For model-based policy search (from which our approach is a particular case), a key requirement of the dynamic model is its ability to produce good long-term predictions. Many approaches optimize a one-step-ahead model, for example by maximizing the likelihood of $p(\st_{t+1}|\st_t, \ic_t)$. However, due to modelling errors, false assumptions on the model and noisy or partial observation model, this approach tends to produce brittle predictions which diverge quickly from the real system \cite{billings2013nonlinear}. Robust approaches such as \cite{doerr2018probabilistic} optimize the likelihood of full sequences of observations $p(\obs_1,\dots, \obs_T)$ marginalized on the sequence of latent states $\{\st_1, \dots, \st_T\}$. Our approach optimizes the same objective but in the generative adversarial framework, which does not require to model explicitly the marginal distribution of observation. It makes very little assumptions on the dynamic, observation and policy models at the expense of a higher computational cost. 

The approach to refine the dynamic parameters is presented in the case of the close-to-optimal discriminator $D_\pds(\traj{})$. 
The proposed process alternates between learning parameters of the policy using Alg.~\ref{alg:adaptvitiy_gamp} and executing this policy on the real system to update the dynamic model. To do so, an additional approximate distribution $\pdatapi$ is introduced. It models the distribution of trajectories executed on the real system with the inferred policy parameters $\ppm$ of the previous step. The distribution of trajectories $\psamples$ sampled with the inferred policy and model of the dynamics is optimized to match this new distribution $\pdatapi$. This time, the gradient is computed with respect to the parameters of the dynamic model $\dpm$.
For increased robustness, it is better to keep multiple dynamic parameters $\{\dpm^{(j)}\}_{j=1}^{L}$ and train them in parallel as an ensemble method, see Alg.~\ref{alg:real_system_gamp}. For example, if the inertia of the robot is not known, the multiple initial dynamic parameters could reflect this. The policy training at the first iteration (before executing on the real system) would be more conservative (high feedback terms) to accommodate this uncertainty. In the case where only one dynamic parameter was used in the policy optimization, the execution on the robot can be disastrous (for example if the inertia matrix was overestimated). 

\begin{algorithm}[h]
	\small
	Compute maximum likelihood of $\pdata$ on the $\nbdata$ demonstrations $\{\hat{\traj{}}^{(i)}\}_{i=1}^\nbdata$ \\
	\For{number of real-system iterations}{
		Start with $L$ initial guesses $\{\dpm^{(j)}\}_{j=1}^{L}$ of system dynamics\\
		
		Update policy with \textbf{Algorithm~\ref{alg:adaptvitiy_gamp}}
		
		Sample $n$ trajectories $\{\hat{\traj{}}^{(i)}_\ppm\}_{i=1}^N$ on the \textbf{real system} given current policy parameters $\ppm$\\
		Compute maximum likelihood of the distribution of new trajectories $\pdatapi$ \\
		\For{number of training iterations}{
			\For{$L$ dynamic models $\{\dpm^{(j)}\}_{j=1}^{L}$}
			{	
				Sample from \eqref{equ:ptraj} $M$ trajectories $\{\traj{(i, j)}\}_{i=1}^M$\\
				Apply (stochastic) maximum likelihood update on $\psamples^{(j)}$ given $\{\traj{(i, j)}\}_{i=1}^M$\\
				Update \textbf{dynamic model parameters} $\dpm^{(j)}$ using the stochastic gradient: 		
				\begin{align}
				\nabla_{\dpm^{(j)}} \frac{1}{M}\sum_{i=1}^{M}-\log\big(\pdatapi(\traj{(i, j)})\big)+
				\log\big(\pdatapi(\traj{(i, j)}) + \psamples^{(j)}(\traj{(i, j)})\big)
				\end{align}
				\vspace{-1em}
			}
				\vspace{-0.5em}
		}
				\vspace{-0.5em}
	}
	\caption{Refining dynamic models with an ensemble method}
	\label{alg:real_system_gamp}
\end{algorithm}


\section{Experiments}
We present two illustrative experiments performed on a 7-DoF Panda robot. Other experiments with quantitative evaluations are performed on synthetic data and presented in Appendix \ref{sec:appendix_experiment}.

\begin{figure}
	\centering
	\begin{subfigure}{.45\textwidth}
		\centering
		\includegraphics[width=.85\linewidth]{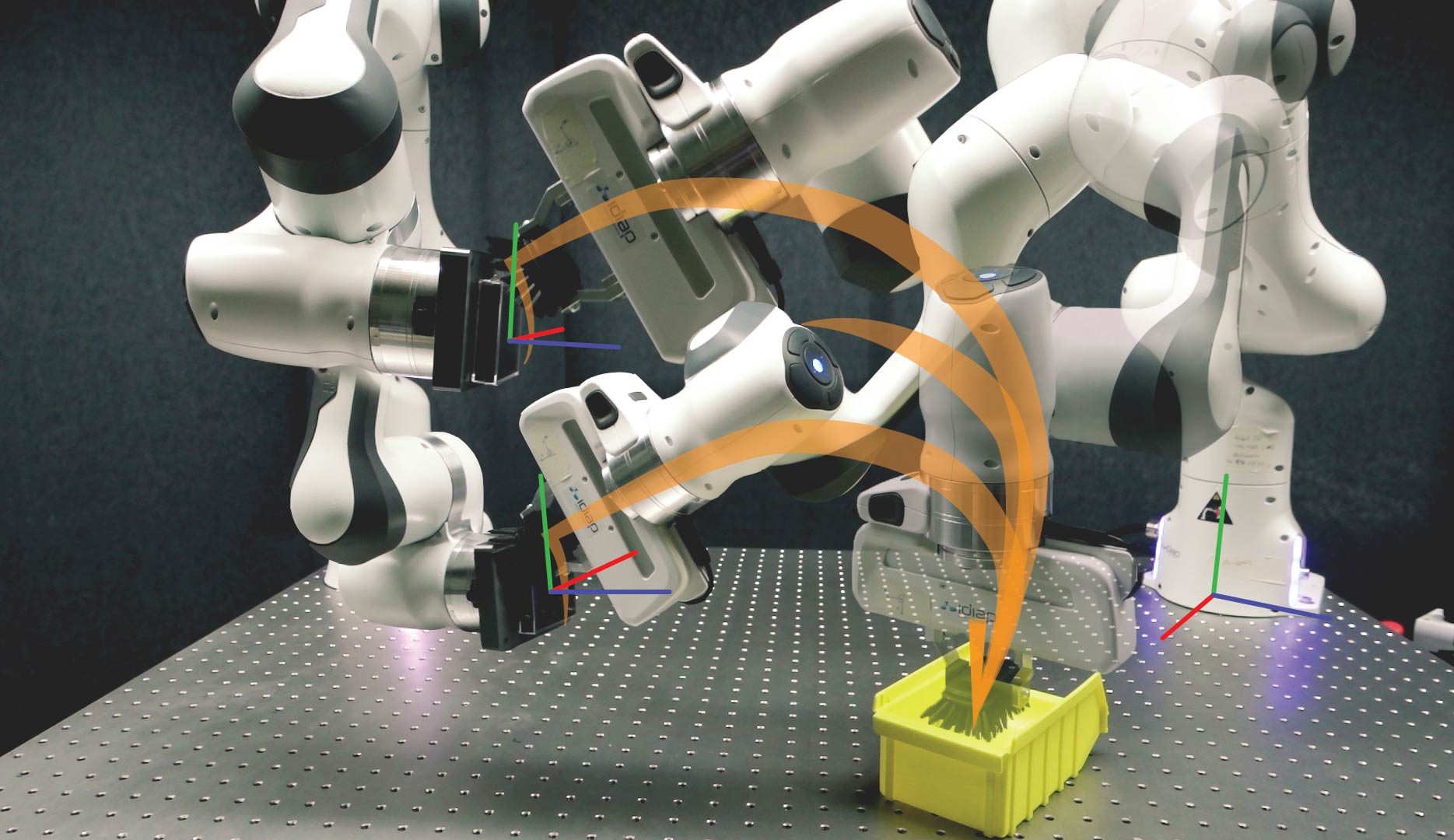}
		\caption{Painting task with adaptation to varying poses.}
		\label{fig:robot_clearning}
	\end{subfigure}
	\begin{subfigure}{.41\textwidth}
		\centering
		\includegraphics[width=.85\linewidth]{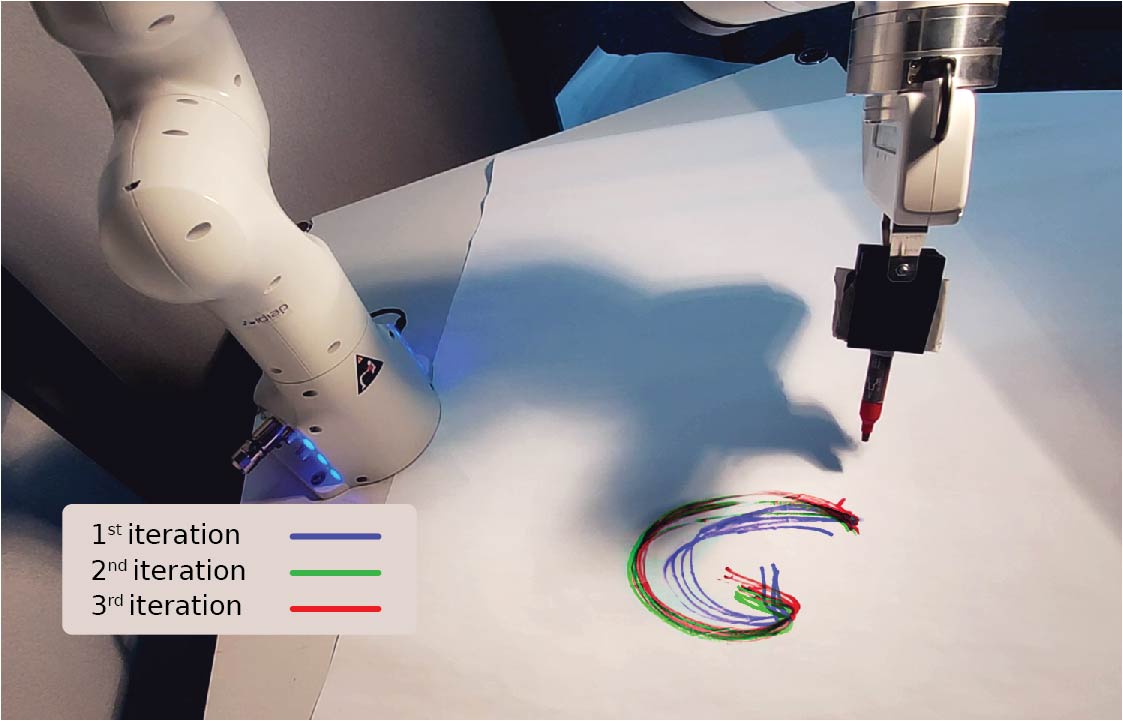}
		\caption{Drawing task in varying environments.}
		\label{fig:robot_drawing_G}
	\end{subfigure}
	\caption{Two illustrative tasks are performed on the robot to demonstrate the adaptation and robustness of the approach. A 7-DoF Panda robot is used in the experiments, controlled with acceleration commands in (a), and with torque commands in (b).}
\end{figure}

\subsection{Acceleration control with adaptation} 
In this experiment, the robot has to paint a box held by another robot. It needs to dip the brush in a paint container that is always at the same place and then wipe the box whose orientation and position can vary, see Fig.~\ref{fig:robot_clearning}. The dataset consists of $\nbdata=7$ time-aligned demonstrations of $4.5\unit{s}$ with a discretization of time $dt = 0.02 \unit{s}$. In each demonstration, the box has a different pose.
We consider that the control commands are the joint accelerations $\ddq\in \mathbb{R}^7$ and the state $\st  \in \mathbb{R}^{14}$ consists of joint angles and velocities of the robot holding the brush. In this configuration, the dynamic model is thus given as a double integrator.

In order to provide adaptation, the policy and discriminator are defined in joint space and two task spaces. The first task space is the position and orientation of the end-effector in a fixed coordinate system and the second is in a coordinate system attached to the box to paint. In each of these three spaces, a Gaussian feedback controller with time-varying gains, feed-forward terms and covariances are defined. More details are given in Appendix \ref{sec:appendix_policies}, Equation \eqref{equ:feedback_force_policy}. The discriminator consists of a factorized Gaussian distribution in each task space, as in \eqref{equ:q_gaussian_task_space}.

As evaluation, we compare the adaptation capabilities with a ProMP conditioned on the 6-DoF pose of the box. As metric, we compute the Bhattacharyya distance\footnote{This distance is computed by approximating the final distribution with a Gaussian on 10 trajectory samples.} (BD) for the distribution of final position and orientation in the coordinate system of the box between the demonstrations and the reproductions. These final poses are shown in Fig.~\ref{fig:robot_clearning}. Results are reported in Table \ref{tab:evaluations_cleaning}. They first are performed on the 7 different contexts of the demonstrations. In this case, ProMP conditioning gives better results. Generalization is then tested by sampling 20 contexts from the Gaussian distribution of poses in the demonstrations. To analyze the extrapolation capabilities, the standard deviation is multiplied by $\sigma_{c}\in\{1, \sqrt{2}, 2\}$. In every case unseen in the demonstrations, the product of policies generalizes better.

\begin{table}
	\begin{center}
		\small
		\caption{Bhattacharyya distance as quantitative evaluation for the painting task.}
		\vspace{0.4em}
		\begin{tabular}{llllll}
			\toprule
			\textbf{} & \textbf{Training}  & \textbf{Testing $\sigma_{c} =1$} & \textbf{Testing $\sigma_{c} =\sqrt{2}$}& \textbf{Testing $\sigma_{c} =2$} \\
			\midrule   
			ProMP conditioning \cite{paraschos2013probabilistic} & $\textbf{0.35} \pm \textbf{0.13}$ & $1.72 \pm 2.29$ & $8.18\pm 15.86 $ & $13.21 \pm 22.66$\\
			GAMP & $0.60 \pm 0.23$ & $\textbf{0.79} \pm \textbf{0.81}$ & $\textbf{1.20} \pm \textbf{0.84}$ & $\textbf{7.02} \pm \textbf{17.78}$\\
			\bottomrule
			\label{tab:evaluations_cleaning}
		\end{tabular}
	\end{center}
\end{table}

\subsection{Force control and dynamic model learning}

In this experiment, we reproduce 2D handwritten letters from \cite{calinon2016tutorial} in different environments. For each letter, $\nbdata=13$ time-aligned demonstrations of $T=200$ timesteps are given. With a discretization of time $dt = 0.01~\mathrm{s}$, the trajectories last $2 \unit{s}$. The policy learned is then run at $1000\unit{hz}$ on the robot. A third dimension is added to the letters as a fixed height. We alternate between optimizing the policy and refining the model of the system, as proposed in Alg.~\ref{alg:real_system_gamp}. We consider that the control commands are the forces $\force\in \mathbb{R}^3$ applied at the end-effector as $\trq = \bm{J}^\trsp \force$. The state $\st  \in \mathbb{R}^6$ is the position and velocity of the end-effector. In this experimental setup, the configuration of the robot is considered as a hidden variable that influences the dynamics. This uncertainty has to be learned by the identification of the dynamic parameters and the policy robust to the unknown configuration.

A time-dependent feedback controller is used as in the first experiment. The dynamics are learned by an ensemble method. We consider that the system is a mass of $3\unit{kg}$ on which a non-linear state-dependent perturbation is added. This non-linear term is modeled as a MLP with two hidden layers of 20 units, $tanh$ activation and the last layer linear. At initialization, the neural networks generate perturbations of a standard deviation of $5\unit{N}$. This initialization is important: if the true system is in the distribution of systems defined by the initialization of the $L$ models, then the first policy executed on the robot will be already quite good. We demonstrate this by performing the same task with a unique neural network instead of an ensemble.

\begin{figure}
	\begin{subfigure}{.73\textwidth}
		\centering
		\includegraphics[width=1.\linewidth]{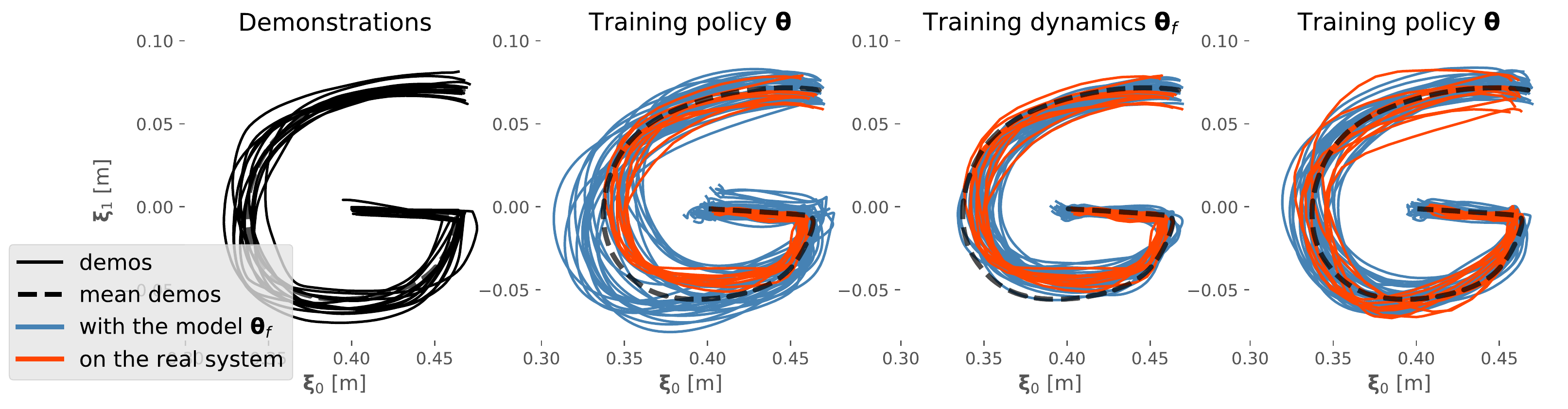}
		\caption{Iterations of Alg.~\ref{alg:real_system_gamp} using an ensemble method to learn the dynamics. 
		}
		\label{fig:robot_dynamics_ensemble_G}
	\end{subfigure}
\hspace{0.01em}
	\begin{subfigure}{.25\textwidth}
		\centering
		
		\includegraphics[width=1.\linewidth]{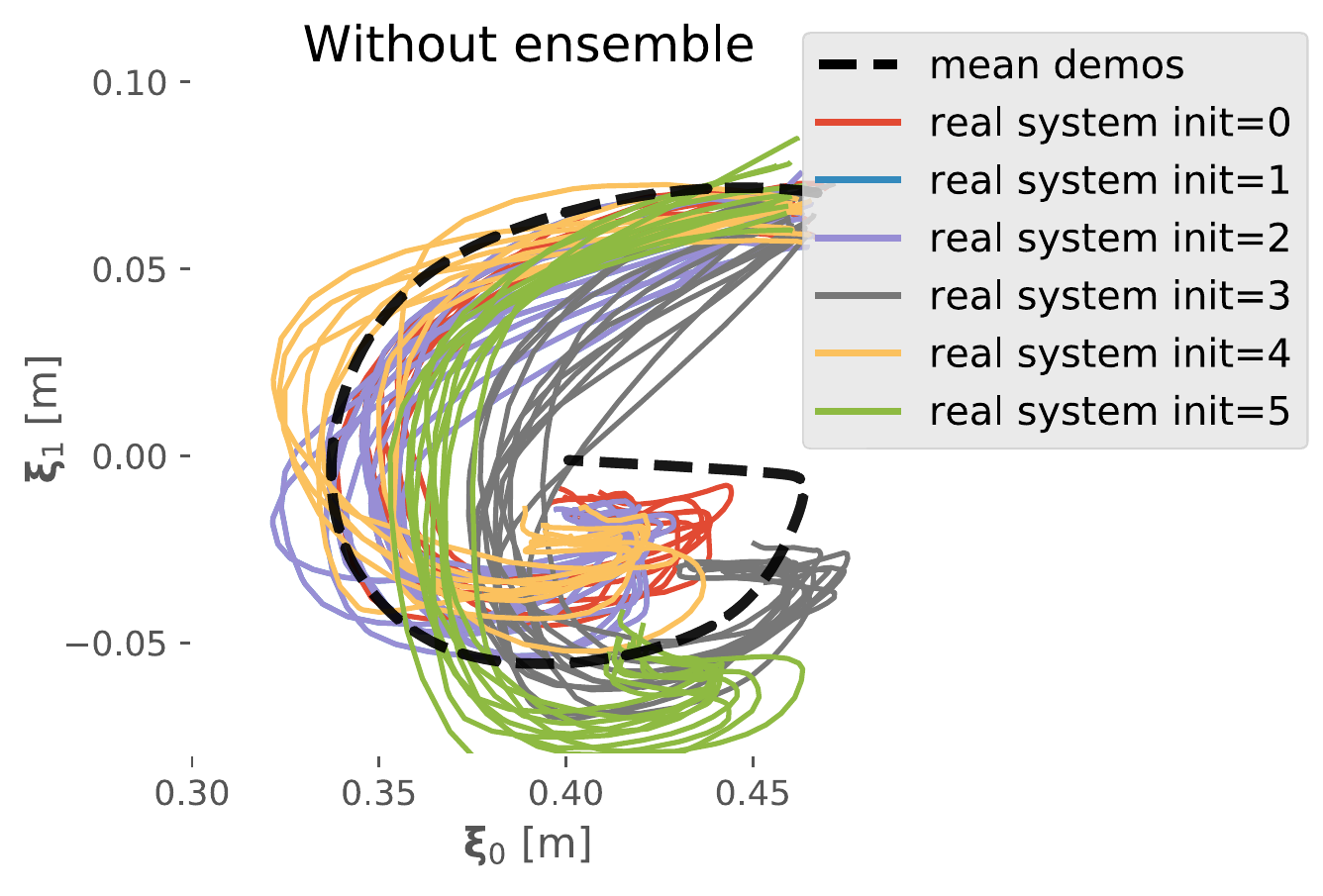}
		\caption{First execution without the ensemble method. 
		}
		\label{fig:robot_dynamics_G}
	
\end{subfigure}
\vspace{-0.4em}
	\caption{
		Learning to reproduce the distribution of \textquotedblleft G\textquotedblright\ letter on a 7-DoF Panda robot.
	}
\end{figure}

After initialization of the dynamic parameters, a robust policy is learned for $10\unit{s}$. This policy is run on the robot for $M=10$ times, by starting at a random initial state of the demonstrations, and with a random configuration. 
With the initial guesses about the system, the first computed policy already leads to a very similar distribution, see Fig.~\ref{fig:robot_dynamics_ensemble_G} (second column).
The $L$ dynamic parameters are optimized in parallel for $10\unit{s}$ given these new trajectories. The trajectories of the generator now match the trajectories on the true system (third column).
The whole process can be repeated until convergence. In this experiment, the policy has been updated only once more for $10s$ and tested on the robot with good results (fourth column). 

As a comparison, Fig.~\ref{fig:robot_dynamics_G} shows the execution of the first policy on the system when no ensemble method are used. Several sets of trajectories are displayed corresponding to different initializations of $\dpm$. In this case, the trajectories are worse than the ensemble method because the epistemic uncertainty is not taken into account, which results in an overconfidence on the dynamic model. As a comparison, the first policy computed using the ensemble method (Fig.~\ref{fig:robot_dynamics_ensemble_G}) has higher feedback gains, resulting in a lower sensitivity to uncertainties.

\begin{figure}
	\centering
	\includegraphics[width=.9\linewidth]{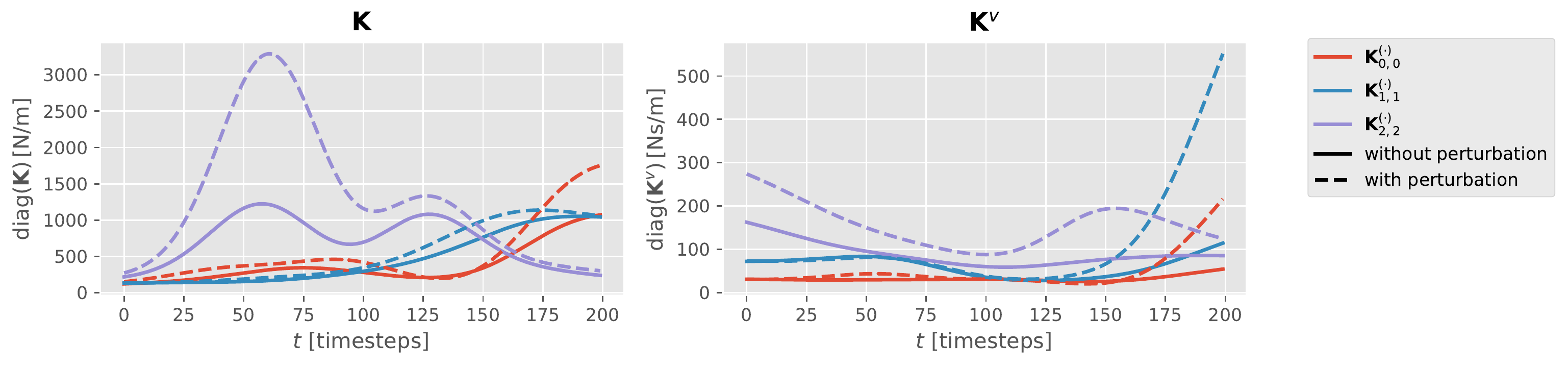}
	\caption{Increase of feedback gains resulting from the identification of external perturbations. \emph{Left:} Diagonal values of the time-dependent proportional gains $\bm{K}(t)\in \mathbb{R}^{3\times3}$. 
	\emph{Right:} Diagonal values of the derivative gains $\bm{K}^v(t)\in \mathbb{R}^{3\times3}$.}
	\label{fig:robot_dynamics_ensemble_G_push}
\end{figure}

These first sets of trajectories were produced without any other perturbations than the unknown configuration. To assess for the generality of the method, we tested the process in three different environments.
In the first, a user applies short (around $0.2\unit{s}$) perturbations of around $10\unit{N}$ all along the trajectories and in every direction. After an update of the dynamic model, a higher stochasticity of the environment is inferred. The following update of the policy results in higher feedback terms (see Fig.~\ref{fig:robot_dynamics_ensemble_G_push}). 

\begin{figure}
	\centering
	\includegraphics[width=1.\linewidth]{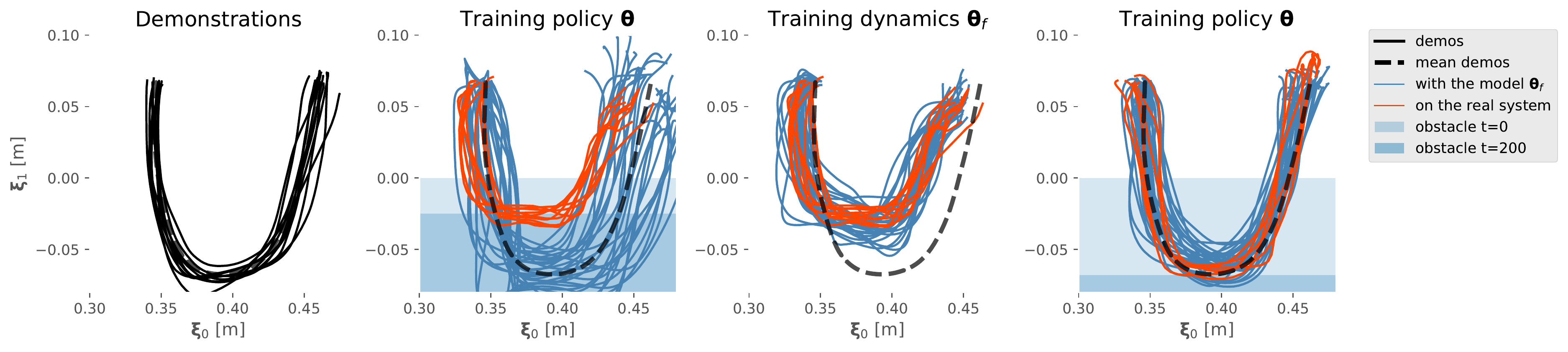}
	\caption{Iterations of Alg.~\ref{alg:real_system_gamp} for reproducing a distribution of \textquotedblleft U\textquotedblright\ shaped trajectories with an obstacle that should be pushed. 
}
	\label{fig:robot_dynamics_ensemble_U} 
\end{figure}

In a second environment, we tested if the approach is able to learn that an obstacle is on its path, which should be pushed. This time, the letter \textquotedblleft U\textquotedblright\ was chosen and a moving plastic block of around $1 \unit{kg}$ put on the table to block the lower part of the letter. 
Iterations of Alg.~\ref{alg:real_system_gamp} are shown for this environment in Fig.~\ref{fig:robot_dynamics_ensemble_U}. The first trajectories executed on the robot are truncated (second column). The friction between the end-effector of the robot and the obstacle also prevents motion along $\st_0$. After updating the dynamics model with $M=10$ rollouts, the prediction of the generator matches the real system (third column). The following update of the policy leads to a much better reproduction of the distribution (fourth column).

In the third case, the robot needs to draw the letter on a paper. A pen was placed in the gripper of the robot and the end-effector redefined as the tip of the pen. The policy was constrained to apply a constant force of $8\unit{N}$ on the paper, as this information was not in the dataset. The contact of the table was explicitly modeled in the dynamics model as a spring-damper system with high gains. The ensemble method still had to learn the additional friction induced by the tip of the pen on the paper. This dynamics was harder to train and the system needed three iterations of the whole process instead of one in the previous experiments. The trajectories of these iterations are shown in Fig.~\ref{fig:robot_drawing_G}.

\section{Conclusion}
The generative adversarial framework is promising for learning movement primitives. It can bring together numerous classical techniques from LfD with the computation power and flexibility of modern machine learning architecture \cite{abadi2016tensorflow}. The approach is easy to be adapted to a wide range of problems. A practitioner vaguely familiar with machine learning is only required to define a function for the dynamic system, one for the policy and possibly multiple relevant task spaces. Future work will focus on learning more complex policies and dynamics models, also from raw pixel observations. More efficient model-based policy search methods should also be incorporated in the framework, to cope with longer horizon problems. 

\clearpage

\acknowledgments{The research leading to these results has received funding from the European Commission's Horizon 2020 Programme through the MEMMO Project (Memory of Motion, \href{http://www.memmo-project.eu/}{http://www.memmo-project.eu/}, grant agreement 780684) and CoLLaboratE project (\href{https://collaborate-project.eu/}{https://collaborate-project.eu/}, grant agreement 820767).}


\bibliography{bib_total}

\appendix

\section{Robotic policies/controllers}
\label{sec:appendix_policies}
In this appendix, we propose several convenient parameterizations of policies that can be used in our framework. As proposed in Sec.~\ref{subsec:pogp}, the policies used in this work are defined in $P$ task spaces $\trans{}{p} : \mathbb{R}^d \rightarrow  \mathbb{R}^{k_p}$. We denote $\x_p= \trans{}{p}(\q)$ the value in task space $p$ and $\J_p =\partial\trans{}{p}/\partial \q$ its Jacobian. The velocity $\dx_p$ and acceleration $\ddx_p$ in the task space are
\begin{align}
\dx_p &= \J_p(\q) \dq,\\
\ddx_p &= \J_p(\q) \ddq + \dot{\J_p}(\q) \dq \approx  \J_p(\q) \ddq.
\end{align}
The relation between the joint torque $\trq$ and the generalized force is 
\begin{align}
\J^{\trsp}_p(\q)\, \force_p = \trq.
\end{align}
These relations are used to define an equivalence between variables used in the above and the different control strategies. The equivalences are reported in Table \ref{tab:equivalences} for different control strategies. For velocity and acceleration control, $\hat{\dq}$ and $\hat{\ddq}$ are reference values that are tracked by lower-level controller, such as inverse dynamics \cite{nakanishi2008operational}.

\begin{table}
	\begin{center}
		\caption{Equivalences between abstract state $\st$, control command $\ic$, task-spaces transform and robotic variables.}
		\begin{tabular}{lccc} 
			\toprule Control strategy & Velocity & Acceleration & Force \\ 
			\bottomrule 
			\midrule 
			State $\st$ & $\q$ & $\begin{bmatrix} \q\\ \dq \end{bmatrix}$ & $\begin{bmatrix} \q\\ \dq \end{bmatrix}$\\ 
			\midrule 
			Control command $\ic$ & $\hat{\dq}$ & $\hat{\ddq}$ & $\trq$\\ 
			\midrule 
			Transform $\transs{}{p}$ & $\trans{}{p}(\q)$ & $\begin{bmatrix}
			\trans{}{p}(\q)\\
			\J_p(\q)\dq
			\end{bmatrix}$ & $\begin{bmatrix}
			\trans{}{p}(\q)\\
			\J_p(\q)\dq
			\end{bmatrix}$\\ 
			\midrule 
			Transform $\transu{}{p}$ & $\J_p(\q)\hat{\dq}$ & $\J_p(\q)\hat{\ddq}$ &  ${\J_p^\trsp}^\dagger(\q)\trq$ \\
			\bottomrule
			\label{tab:equivalences}
		\end{tabular}
	\end{center}
\end{table}

These relations do not need to be exact. They are just parameterizations of the policy which give a better structure to the problem to facilitate the training phase and increase generalization capabilities. Simplifications, such as dropping $\dot{\J_p}(\q)$ for acceleration control, can be done to speed up the computation of the stochastic gradient while training. 

The policies can be parametrized in different ways. For time-dependent policies, a solution is a feedback controller with time-varying gains and feed-forward terms. 
These controllers are very usual in LfD \cite{paraschos2013probabilistic, calinon2016tutorial} and are also solutions of linear-quadratic tracking problems \cite{bohner2011linear}.
They can be used both for velocity control
\begin{align}
\mup_p^t(\transs{t}{p}(\st)) &= -\bm{K}_p(t) \trans{t}{p}(\q) + \tar_p(t),\\
\Sigmap_p^t(\transs{t}{p}(\st)) &= \hat{\Sigmap}_p(t),
\end{align}
or acceleration and force with
\begin{align}
\mup_p^t(\transs{t}{p}(\st)) &= -\bm{K}_p(t)  \trans{t}{p}(\q) - \bm{K}_p^v(t)\J_p(\q)\dq + \tar_p(t).
\label{equ:feedback_force_policy}
\end{align}
Continuous values for the parameters depending on time $t$ can be induced by linear basis functions or simple multilayer perceptrons (MLP). Gains $\bm{K}$ can be parametrized in several restrictive ways depending on the assumptions on the system. It can help at stabilizing and speeding up the training phase as well as at providing more safety on the robot.   

Time-independent policy can be defined with MLPs that output the parameters of the Gaussian policy
\begin{align}
\mup_p^t(\transs{t}{p}(\st)) = \nn_\mup\big(\trans{t}{p}(\q)\big), \quad
\Sigmap_p^t(\transs{t}{p}(\st)) = \nn_\Sigmap\big(\trans{t}{p}(\q)\big).
\label{equ:mlp_policy}
\end{align}
Covariance matrices can be parametrized by their Cholesky decomposition or using the matrix exponential of another symmetric matrix. 
Time-independent policies have also been modeled by conditioning in Gaussian mixture models \cite{khansari2011learning, pignat2019bayesian}. This latter approach is extremely fast to train from data but its gradient is not well-behaved for optimization.

\section{Additional experiments with synthetic data}
In this appendix, we propose additional experiments with synthetic data and more exhaustive quantitative evaluations. In the first experiment, we compare the matching of distributions under simulated perturbations. In the second experiment, we learn a context-dependent time-independent policy and test its robustness and generalization capabilities.
\label{sec:appendix_experiment}
\subsection{Time-dependent policy} 
\begin{figure}
	\centering
	\includegraphics[width=.8\linewidth]{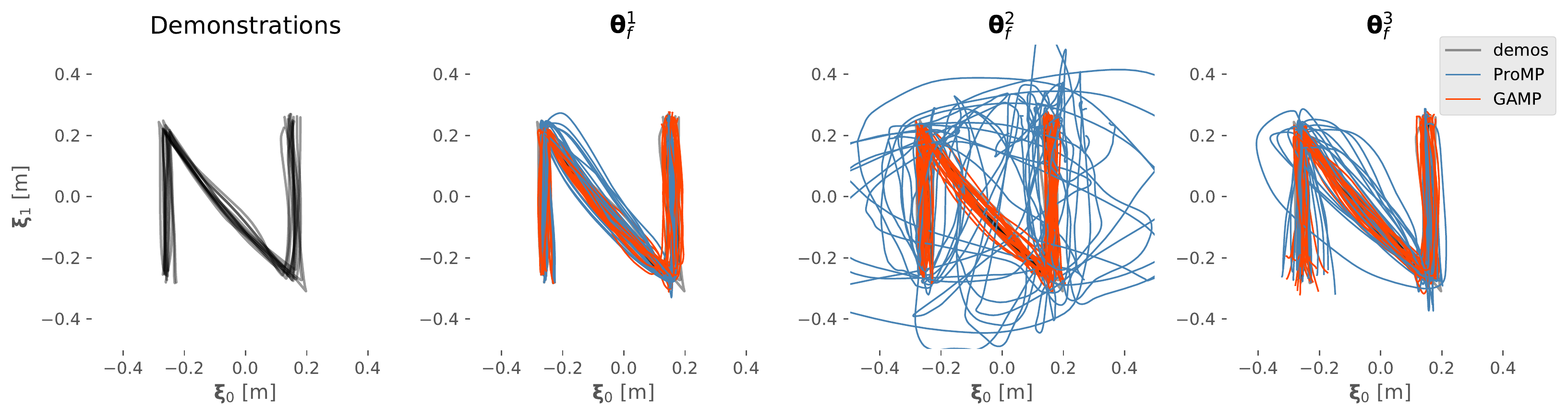}
	\caption{Demonstrations and reproductions using ProMP and GAMP for different stochasticity of the environment. The same controllers are used in the 3 situations.}
	\label{fig:dataset_repros_letter_feedback}
\end{figure}

In this first experiment with synthetic systems, we consider a simulated 2D unit mass system with a discretization of time $dt=0.01 \unit{s}$. The state $\st\in \mathbb{R}^4$ is composed of its position and velocity, and the control command $\ic \in \mathbb{R}^2$ the force.
The dataset are letters from the alphabet \cite{calinon2016tutorial}. For each letter, $\nbdata=13$ time-aligned demonstrations of $T=200$ timesteps are given. Demonstrations are shown in Fig.~\ref{fig:dataset_repros_letter_feedback}-(\textit{left}) for letter \textquotedblleft N\textquotedblright . 
A time-dependent feedback Gaussian policy as in \eqref{equ:feedback_force_policy} is used. The gain matrices and feed-forward terms are parametrized with time-dependent basis functions. Gains are parametrized in several ways, which has more influence on training time than on the final results. For the evaluations, $\bm{K}_p(t)$ and $\bm{K}_p^v(t)$ were chosen as diagonal matrices with positives elements. 
In this first experiment, the policy is not a POGP as it is defined only in the original state space.
The approximate distributions used for the discriminator are factorized Gaussians as in \eqref{equ:q_gaussian}. Each letter was trained for $10\mathrm{s}$. The approach presented in Sec.~\ref{sec:stochastic} was used on a distribution of dynamic parameters $p(\dpm)$ which include different values of Gaussian perturbations in force and on the initial state.

We compare our approach with ProMP \cite{paraschos2013probabilistic} and hybrid approaches that learn a time-dependent distribution of states with either Gaussian mixture regression (GMR + LQT) \cite{calinon2009statistical} or hidden Markov models (HMM + LQT) \cite{calinon2016tutorial} and use linear quadratic tracker to regenerate continuous trajectories. 

Given the controller computed for each model, we evaluated rollouts in 3 situations. In the first case, $\dpm^{(1)}$, the system is deterministic. In the second case, $\dpm^{(2)}$, uncorrelated Gaussian perturbations in force of standard deviation of $10\unit{N}$ are injected. In the third case, $\dpm^{(3)}$ Gaussian noise on the initial position of standard deviation of $0.035\unit{m}$ is applied. Two metrics are used to compare the demonstrations with the synthesized samples. The first one evaluates if the mean motion is well reproduced. For each letter, the mean squared error (MSE) is computed between the mean trajectories (position and velocity only) over the $\nbdata=13$ demonstrations and the mean over 20 samples from the model. The second metric evaluates if the full distribution is well reproduced. The Bhattacharyya distance (BD) is computed over a Gaussian approximation of the distribution of demonstrations and samples.

The results are reported in Table \ref{tab:comp_gamp_promp_lqt} for each case of dynamics, each model and the two metrics. The mean value and standard deviation of these metrics over the whole alphabet is given.
Demonstrations and synthesized samples from ProMP and GAMP are shown in Fig.~\ref{fig:dataset_repros_letter_feedback} for each stochasticity.
ProMP and GAMP have similar results in terms of MSE and BD in case of no stochasticity $\dpm^{(1)}$. The ProMP controller derived in \cite{paraschos2013probabilistic} assumes known dynamics and stochasticity in order to match the distribution of demonstrations. Our approach can generate a controller that matches the distribution, for a distribution of stochasticity of the environment. A more robust controller for ProMP can be derived in our framework by using $\pdata$ and $\psamples$ as ProMP distributions. The approach using GMR + LQT is robust to perturbations and performs well in terms of MSE. However, the distribution is not matched very well. The velocity of the rollouts have the same variance as the demonstrations but the positions tend to shrink on the mean trajectory. This is due to the assumption of independence between two consecutive states that are ignored in GMR and HMM and that are later restored with LQT. The proper way to generate the matching distribution would be to use IOC with LQT \cite{finn2016connection} or trajectory HMM \cite{zen2007reformulating}.

\begin{table} 
	\scriptsize
	\begin{center}
		\caption{Mean squared error and Bhattacharyya distance between demonstrations and samples of different models. The samples are generated with three different dynamic parameters.}
		\vspace{0.4em}
		\begin{tabular}{lllllll}
			\toprule
			\textbf{Metrics} & \multicolumn{3}{l}{\textbf{Mean squared error}}  & \multicolumn{3}{l}{\textbf{Bhattacharyya distance}}   \\
			Environment $\dpm$ & $\dpm^{(1)}$ & $\dpm^{(2)}$ & $\dpm^{(3)}$ & $\dpm^{(1)}$ & $\dpm^{(2)}$ & $\dpm^{(3)}$\\
			\midrule    
			GAMP (ours)   & $\textbf{0.18} \pm \textbf{0.06}$        & $\textbf{0.19} \pm \textbf{0.08}$    & $\textbf{0.23} \pm \textbf{0.11} $ &
			$\textbf{0.13} \pm \textbf{0.04}$        & $\textbf{0.13} \pm \textbf{0.04}$    & $\textbf{0.11} \pm \textbf{0.04}$ \\
			ProMP \cite{paraschos2013probabilistic}    & $0.38 \pm 0.27$        & $1.15 \pm 0.55$    & $0.60 \pm 0.42$ &
			$0.24 \pm 0.10$        & $0.90 \pm 0.30$    & $0.29 \pm 0.10$ \\
			GMR + LQT \cite{calinon2009statistical}  & $0.31 \pm 0.14$        & $0.34 \pm 0.13$    & $0.31 \pm 0.13$ &
			$0.40 \pm 0.07$        & $0.30 \pm 0.06$    & $0.40 \pm 0.07$ \\
			HMM + LQT \cite{calinon2016tutorial}  & $1.58 \pm 0.48$        & $1.20 \pm 0.47$    & $1.20 \pm 0.48$ &
			$0.76 \pm 0.21$        & $0.58 \pm 0.21$    & $0.75 \pm 0.21$ \\
			\bottomrule
			\label{tab:comp_gamp_promp_lqt}
		\end{tabular}
	\end{center}
\end{table}

\subsection{Time-independent policy in two task spaces} 

\begin{figure}
	\begin{subfigure}{.5\textwidth}
		\centering
		\includegraphics[width=1.\linewidth, page=1]{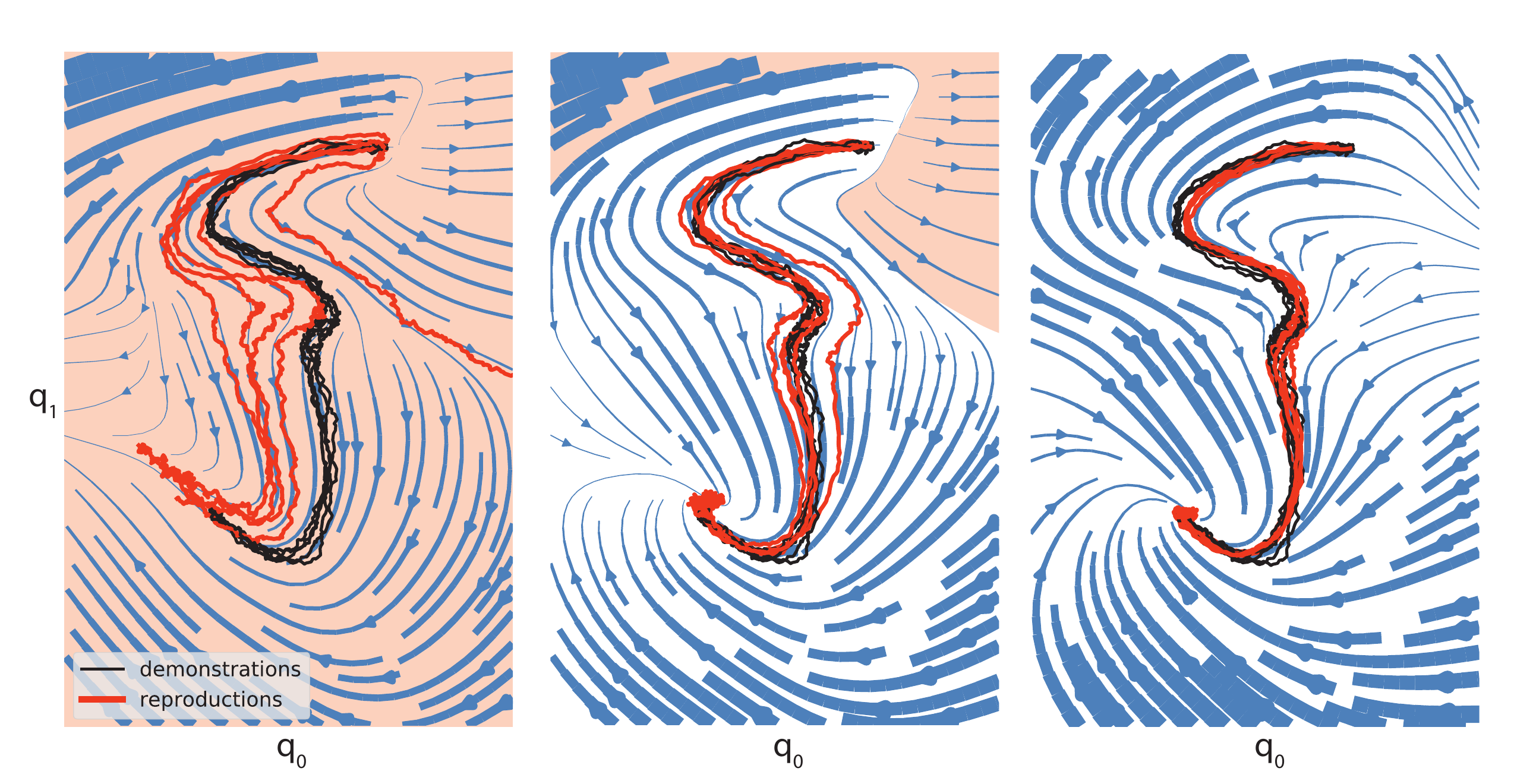}
		\caption{Demonstrations and samples with several learning strategies. The area leading to divergence are highlighted. (\textit{left}) The policy is trained only by maximizing its own likelihood, as in policy imitation. (\textit{middle}) The policy is trained with the method presented in this paper, to maximize the likelihood of the trajectories. (\textit{right}) Additional perturbations are added in the initial state of each chunk, leading to a more robust policy.}
		\label{fig:dataset_repro_tp_nn_imitation}
	\end{subfigure}
	\hspace{0.4em}
	\begin{subfigure}{.5\textwidth}
		\centering
		\includegraphics[width=1.\linewidth, page=2]{img/gamp.pdf}
		\caption{Demonstrations and samples in several situations. The transformation of the moving object is indicated with a coordinate system. The flow field displays only the mean value of the policy. }
		\label{fig:dataset_repro_tp_nn}
	\end{subfigure}
	\caption{Illustration of time-independent policies as flow fields.}
	\label{fig:repro_tp_nn}
\end{figure}

In the second experiment, we consider a time-independent policy that should adapt to a moving object, as shown in Fig.~\ref{fig:dataset_repro_tp_nn}. The system considered is a simple integrator with possible perturbations.
The dataset consists of a smooth blend between letters \textquotedblleft S\textquotedblright\ and \textquotedblleft J\textquotedblright. The letter \textquotedblleft J\textquotedblright\ should move according to an object. Ten different positions and orientations of the object are given, and for each of which $\nbdata=10$ demonstrations of $T=400$ timesteps are performed. The dataset was randomly split into 5 situations to train and 5 to test the generalization.
The policy is the product of two time-independent Gaussian policies given by an MLP as in \eqref{equ:mlp_policy}. These two policies are defined on a different task space: the first one on a fixed task space, and the second one, projected in the coordinate system of the moving object. The MLPs have both 2 hidden layers of 150 units with $tanh$ activation and output a state-dependent Gaussian with a full covariance. The covariance is parametrized by its matrix logarithm. Even if the demonstrations are full and aligned, we discard this information for training, and randomly split them in small chunks. 
The approximate distributions used for the discriminator are Gaussian mixture models with $K=20$ as in \eqref{equ:q_mixture} in each task space. Before each update of policy parameters $\ppm$, 10 steps of EM are performed with 1000 points each. Every 50 steps of policy parameters update, the mixture models are reinitialized with k-means to avoid local minima.
The policy parameters are initialized by maximum likelihood of the policy density on pairs of $\{\st, \ic\}$ for $5\unit{s}$ of stochastic gradient descent. This initialization corresponds to a policy imitation objective, which is known to produce brittle policy \cite{ross2011reduction}. Fig.~\ref{fig:dataset_repro_tp_nn_imitation}-\textit{(left)} shows the policy after initialization and the dangers of drifting away from the training data. The models are further trained for $15 \unit{s}$ in the generative adversarial network. Two alternatives are considered. In the first (GAMP), the system is assumed to be deterministic during training. In the second (GAMP + noise), small perturbations in the initial state of the chunks are simulated. The policy is then trained to look like the demonstrations, even with noise, which results in more robustness. The differences between the policy learned in these two cases are shown in Fig.~\ref{fig:dataset_repro_tp_nn_imitation}-\textit{(middle and right)}.
We also consider another policy, where the adaptation to the moving object is given by the neural network instead of the usage of multiple task spaces. In this case, the MLP defining the Gaussian policy has an additional input. It is the position and vectorized rotation matrix of the object.

As an evaluation, we produce full rollouts from the initial states of the demonstrations. We compute the mean absolute error (MAE) over the position of the whole rollout and the closest demonstration. The mean value and its standard deviation over the training and testing situations are given in Table \ref{tab:evaluations_tp_nn}. 
\begin{table}
	\begin{center}
		\caption{Mean absolute error (MAE) between the whole rollout and the closest demonstrations for situations in the training and testing set. The red colour indicates huge errors because of divergence.}
		\vspace{0.4em}
		\begin{tabular}{lllllll}
			\toprule
			&\textbf{} & \textbf{Training}  & \textbf{Testing}  \\
			\midrule   
			\multicolumn{3}{l}{\textit{Task spaces adaptation}}\\
			&GAMP + noise  & $\textbf{0.016} \pm \textbf{0.003}$        & $\textbf{0.025} \pm \textbf{0.009} $\\
			&GAMP  & $0.019 \pm 0.008$        & $0.075 \pm 0.13$     \\
			&Imitation  & $0.251 \pm 0.346$        & $1.629 \pm 2.124$    \\
			\midrule   
			\multicolumn{3}{l}{\textit{MLP adaptation}}\\
			&GAMP + noise & $0.017 \pm 0.002$        & $0.078 \pm 0.014$    \\
			&GAMP  & $0.025 \pm 0.008$        & $0.086 \pm 0.028$     \\
			&Imitation  & \color{red}{$3.2\mathrm{e}{6} \pm 8.7\mathrm{e}{6}$}       &  \color{red}{$1.3\mathrm{e}{6} \pm 6.4\mathrm{e}{6}$}     \\
			\bottomrule
			\label{tab:evaluations_tp_nn}
		\end{tabular}
	\end{center}
\end{table}
The adaptation with the use of task spaces and MLP perform the same on the training set but the former generalizes better. In robotics, many adaptations of movements can be understood easily by projecting them into several coordinate systems. The two approaches can also be combined in the case where the definition of multiple task spaces is not sufficient. The addition of noise in the initial state makes the GAMP more robust. They also generalize better to new situations.
When using the imitation cost only, the trajectories diverge, as shown in Fig.~\ref{fig:dataset_repro_tp_nn_imitation}-(\textit{left}). In the case of MLP adaptation, they diverge extremely fast, leading to an enormous cost. When using the imitation cost only, the benefits of defining two policies on low-dimensional task space instead of a unique policy with an additional vector of inputs are important. The fusion of two robust policies will tend to be more robust than a policy that can change completely for each new vector defining the situation. 

\begin{figure}
	\centering
	\includegraphics[width=.8\linewidth, page=1]{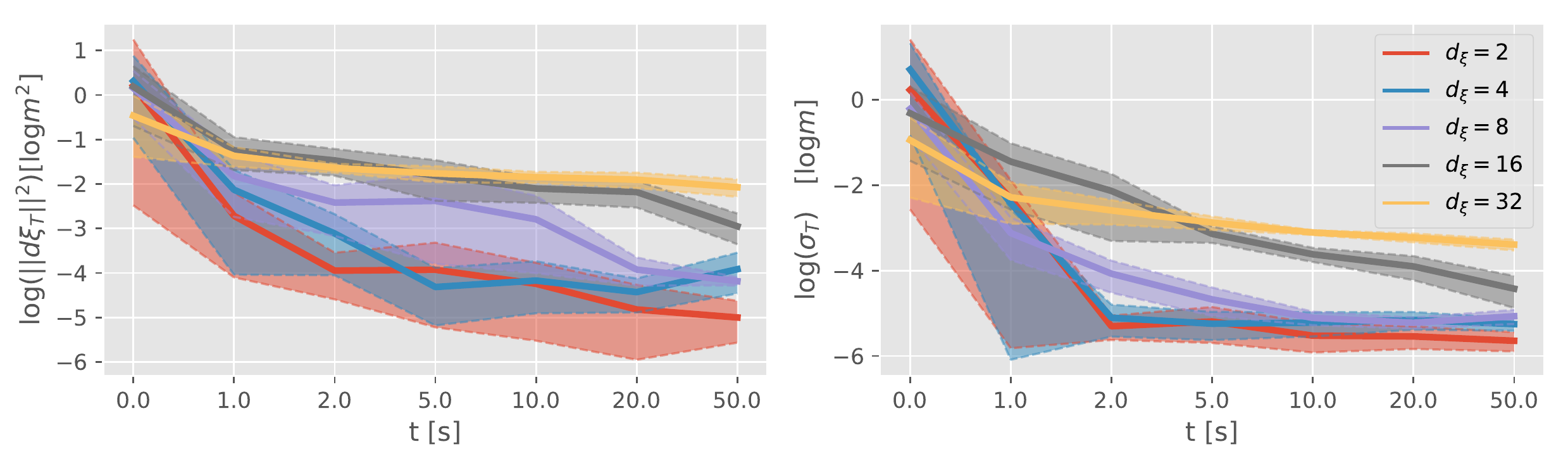}
	\caption{Evaluations of the robustness for different dimensionality of state $d_\st \in \{2, 4, 8, 16, 32\}$. (\textit{left}) Log mean distance between the final state of the demonstrations and the 10 rollouts.  (\textit{right}) Log mean standard deviation of the final state of the 10 rollouts.}
	\label{fig:evaluations_high_dim_nn}
\end{figure}

The problem of learning robust policy is very difficult \cite{ross2011reduction}, \cite{khansari2011learning}. Our approach gives no guarantees that the system cannot diverge, but the cost of mimicking the distribution of trajectories greatly increases the robustness. 
By injecting noise and training with stochastic gradient descent for a sufficient amount of time, the system is expected not to diverge. We performed an additional test for showing that this also applies to higher-dimensional systems, given slightly longer optimization. We created higher-dimensional dataset by randomly concatenating letters up to $\st\in\mathbb{R}^{32}$. Two metrics were used to check the divergence. The first one is the mean distance between the final state of the demonstrations and 10 rollouts.  The second one is the standard deviation of the final state for each rollout. The rollouts were executed for twice the horizon of the demonstrations, to check if the system drifts further and by adding perturbation on the initial state. These two metrics were evaluated for 5 random concatenations of letters for each dimension. The metrics were evaluated just after initialization using the policy imitation cost and several times during $50\unit{s}$ of training. Results are reported in Fig.~\ref{fig:evaluations_high_dim_nn}. They show that, as expected, high-dimensional systems tend to be more difficult to train. However, after a few seconds of optimization, no more trajectories were diverging even for high-dimensional system. They all converged within an area of at worse $0.05\unit{m}$ of standard deviation, while the scale of the workspace is about $1\unit{m}$. 

\end{document}